%% file: main.tex
\documentclass[letterpaper, 10 pt, journal, twoside]{IEEEtran}
\usepackage{ulem}
\usepackage{hyperref}
\usepackage{url}
\usepackage{graphics} 
\usepackage{times} 
\usepackage{booktabs}       
\usepackage{amsfonts}      
\usepackage{nicefrac}      
\usepackage{microtype}    
\usepackage{algorithm,multirow,xcolor}
\usepackage{algorithmic}
\usepackage{mathtools}
\usepackage{graphicx}
\usepackage{cases}
\usepackage{array}
\usepackage{color}
\usepackage{float}
\usepackage{amssymb}
\usepackage{wrapfig}
\usepackage{subfigure}
\usepackage{amsmath}
\usepackage{cite}
\usepackage{soul}

\usepackage{subfigure}
\usepackage{epstopdf}
\usepackage{caption}
\usepackage{threeparttable}

\IEEEoverridecommandlockouts 

\title{\LARGE \bf
Get It For Free: Radar Segmentation without Expert Labels 
and \\ Its Application in Odometry and Localization}

\author{Siru Li$^{1\dagger}$,  Ziyang Hong$^{2\dagger}$, Yushuai Chen$^{1}$, Liang  Hu$^{1*}$ and Jiahu Qin$^{3}$
\thanks{$^{\dagger}$ Equal Contribution, $^{*}$ Corresponding Author (Email:~l.hu@hit.edu.cn)}
\thanks{$^{1}$S. Li, Y. Chen and L.~Hu are with the Department
of Automation, School of Mechanical Engineering and Automation, Harbin Institute of Technology, Shenzhen,
China. $^{2}$Z. Hong is with the School of Data Science, the Chinese University of Hong Kong, Shenzhen, China. $^{3}$ J. Qin is with the Department of Automation, University of Science and Technology of China, Hefei, China}
}

\begin{document}

\maketitle
\thispagestyle{empty}
\pagestyle{empty}

\begin{abstract}

This paper presents a novel weakly supervised semantic segmentation method for radar segmentation, where the existing LiDAR semantic segmentation models are employed to generate semantic labels, which then serve as supervision signals for training a radar semantic segmentation model. The obtained radar semantic segmentation model outperforms LiDAR-based models, providing more consistent and robust segmentation under all-weather conditions, particularly in the snow, rain and fog. 
To mitigate potential errors in LiDAR semantic labels, we design a dedicated refinement scheme that corrects erroneous labels based on structural features and distribution patterns. 
The semantic information generated by our radar segmentation model is used in two downstream tasks, achieving significant performance improvements. In large-scale radar-based localization using OpenStreetMap, it leads to localization error reduction by 20.55\% over prior methods. For the odometry task, it improves translation accuracy by 16.4\% compared to the second-best method, securing the first place in the radar odometry competition at the Radar in Robotics workshop of ICRA 2024, Japan\footnote{https://sites.google.com/view/radar-robotics/competition}. 

\end{abstract}

\begin{IEEEkeywords}
Deep learning methods, SLAM, Semantic segmentation, mmWave radar 
\end{IEEEkeywords}

\section{INTRODUCTION}
The use of cameras and LiDAR for environmental perception has significantly advanced the development of autonomous driving. However, both sensors face challenges in adverse weather conditions such as rain, snow and fog due to occlusion from various particles. In contrast, radar which operates in a lower frequency band within the GHz range offers robust and reliable perception under such conditions. Despite this advantage, radar data generally lacks the precision and quality of camera and LiDAR data due to issues like data sparsity and sensor-specific noise. Consequently, several effective noise reduction methods have been proposed for radar denoising, particularly using convolutional neural networks (CNNs) \cite{weston2019probably,weston2021there,yin2020radar,kung2022radar}.

Semantic information plays a crucial role in complementing visual inputs and has been applied in many advanced semantic visual simultaneous localization and mapping (SLAM) and navigation methods \cite{salas2013slam++,roth2024viplanner}. However, semantic annotations remain absent in most existing radar datasets  \cite{burnett2023boreas,barnes2020oxford}, not to mention downstream tasks semantic radar SLAM. To fill the gap, we propose a weakly supervised learning approach to radar semantic segmentation. Specifically, we generate the semantic labels using a state-of-the-art LiDAR-based semantic segmentation model \cite{lai2023spherical}, and refine these labels through a scheme that corrects erroneous labels based on structural features and distribution patterns. These refined labels serve as cross-modal supervisory signals for training radar semantic segmentation model.  
Compared to manual annotation, this method automatically generates semantic labels and thus is far more efficient and cost-effective, making it highly suitable for large-scale datasets.  Fig. 1 showcases different sensor data and radar semantic segmentation results from our model.

\begin{figure}[t]
    \centering
    \includegraphics[width=\linewidth]{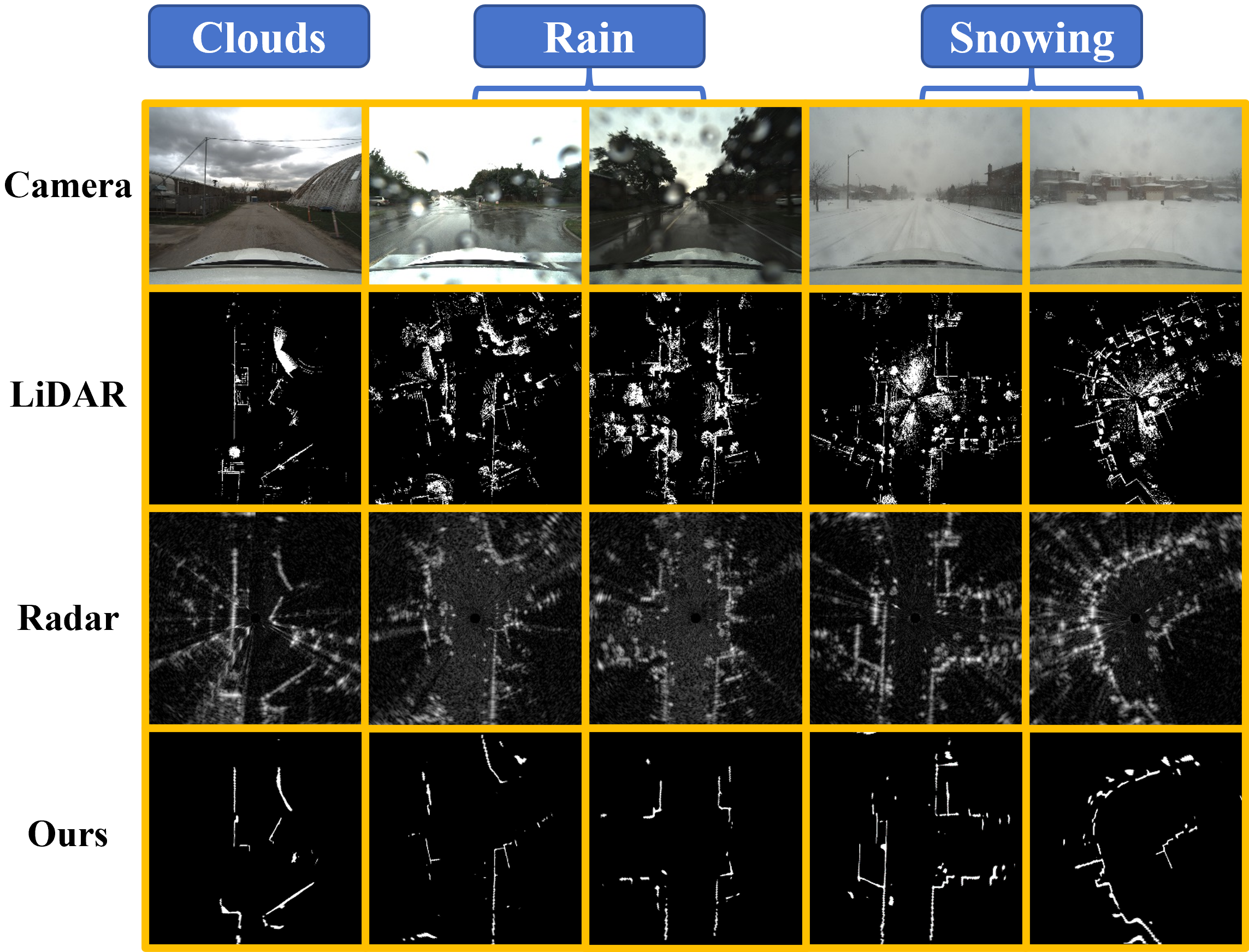}
        \caption{Comparison of the information obtained by the three sensors under three different weather conditions. In rainy and snowy weather, camera images exhibit significant blurring, LiDAR data is heavily obstructed, while the radar remains stable. Meanwhile, our network continues to perform reliable semantic segmentation.}
    \label{generalization}
\end{figure}

\begin{figure*}[ht]
    \centering
    \includegraphics[width=\linewidth]{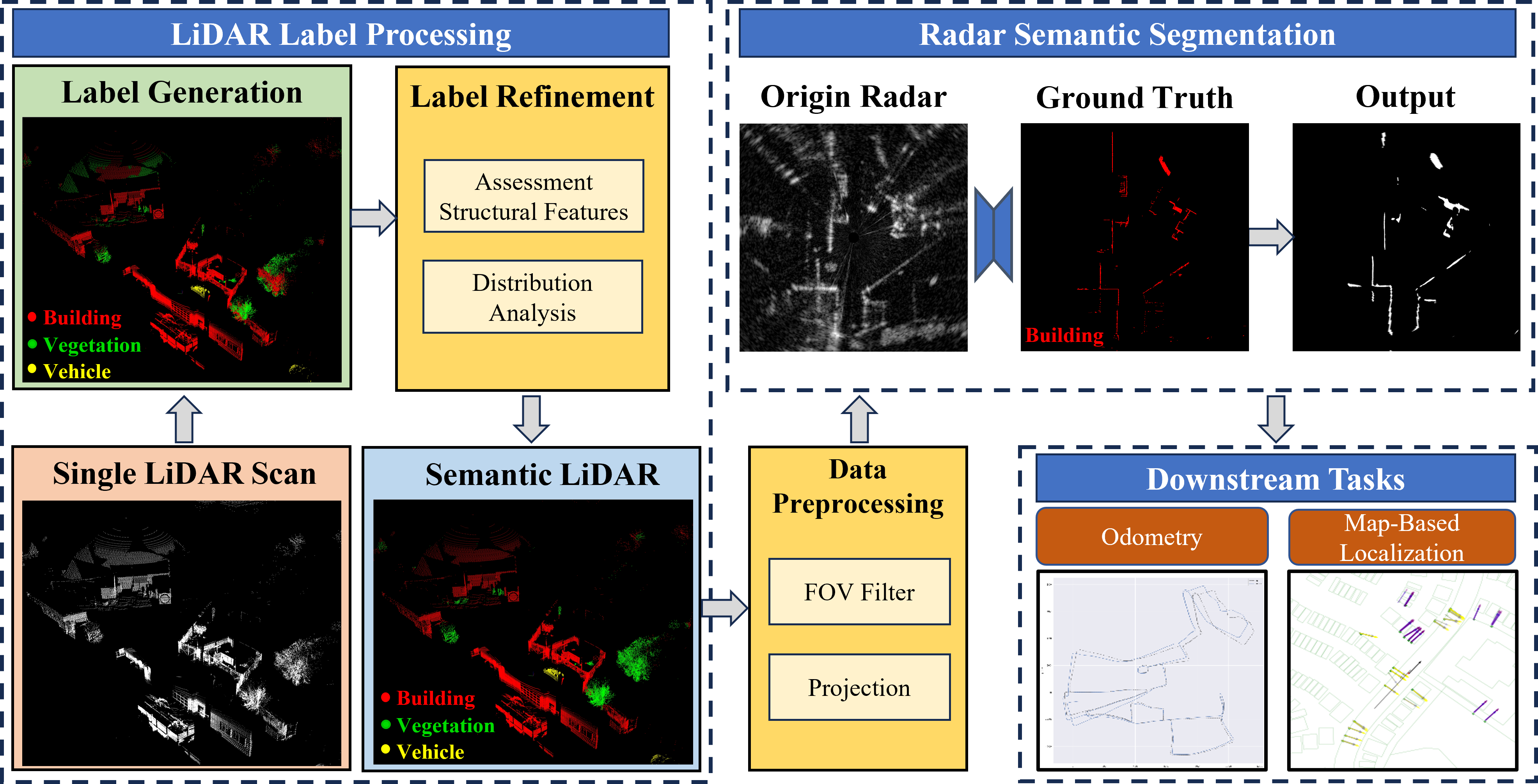}
        \caption{System overview. The system employs another model to generate labels for the LiDAR point cloud. Following label refinement and LiDAR data filtering, the data is projected to create a supervisory signal with specific semantic information. This enables semantic segmentation of radar for downstream tasks.}
    \label{overview}
\end{figure*}
We demonstrate the effectiveness of our radar semantic segmentation model in two downstream tasks: odometry and localization. In our previous work \cite{hong2023large}, we combined radar with OpenStreetMap (OSM) for large-scale localization. However, the absence of semantic information made it difficult to classify points used for computing normal vectors, sometimes even mistaking noise for real objects. By incorporating radar semantic information, our method ensures that normal vector registration focuses specifically on stable objects like buildings.  For odometry estimation tasks, most algorithms are based on the assumption that the environment is static, and moving objects such as cars and pedestrians which are ubiquitous in outdoor environments, if not identified and removed correctly,  could cause incorrect feature matches between frames and hence degrade odometry accuracy. With our method, semantic information allows us to selectively retain stable features from static objects like buildings, improving the accuracy of odometry estimation.

The main contributions are as follows:
\begin{enumerate}
    \item We propose a novel method to refine LiDAR semantic labels over those obtained from the SOTA LiDAR semantic segmentation model based on structural features and distribution patterns. This approach improves the accuracy and reliability of the labels while avoiding the high costs associated with manual annotation;
    \item To the best of our knowledge, we propose the first method to learn a weakly supervised semantic segmentation model for radar that only relies on LiDAR as the supervisory signal;
    \item In downstream odometry and localization tasks, we validate the robustness of our approach across multiple public datasets, demonstrating strong performance even under extreme weather conditions, e.g. heavy rain and snow. 
\end{enumerate}

\section{RELATED WORK}
\subsection{Deep Learning and Radar Perception}
The raw power-distance images generated by scanning radar offer relatively high angular and distance resolution. However, these radar images often suffer from significant noise. Over the past few decades, researchers have employed various approaches to filter this noise, including constant false-alarm rate (CFAR) filtering \cite{rohling1983radar} and static thresholding \cite{kung2021normal}. However, these traditional methods have limited effectiveness in addressing some noise sources such as receiver saturation. With recent advancements in deep learning, many researchers have turned to deep learning methods to train models for filtering radar data, using LiDAR as a supervisory signal \cite{kaul2020rss,yin2020radar}. Moreover, to address the issue of deep learning leading to the loss of some features, the method of sliding window is employed to preserve long-range sensing and penetrating capabilities \cite{kung2022radar}. Additionally, by projecting semantically segmented images onto the point cloud, RSS-Net \cite{kaul2020rss} outputs radar data with semantic information.  Beyond cameras, point cloud data can also be enriched with semantic information. We propose a method that directly generates labels on the point cloud, creating a lidar supervisory signal with semantic information, thereby enabling semantic segmentation of radar. 

\subsection{Spinning Radar for Odometry and Localization}
Spinning radar odometry has become an important focus in robotics and autonomous vehicles, particularly due to its effectiveness in environments where traditional sensors like cameras and LiDAR face limitations. Early methods focused on adapting techniques from visual odometry to radar data, leveraging feature extraction and matching to estimate motion. 
Hong et al. \cite{hong2022radarslam} proposed a dual-process approach in RadarSLAM for odometry estimation. One process performs keyframe matching to estimate the vehicle's pose and produces an initial odometry result, while the other process detects loop closure and uses graph optimization to refine the trajectory, thereby enhancing odometry accuracy. Since spinning radar data does not contain Doppler information, it cannot directly estimate velocity. To address this, Kung et al. utilized an inertial measurement unit (IMU) for velocity estimation and employed weighted Normal Distributions Transform (NDT) for point cloud registration. Additionally, Daniel et al. [11] focused on filtering by selecting the $n$-strongest points in each azimuth direction and computing normal vectors based on adjacent grid points. They preserved intensity information and used these normal vectors for registration, successfully achieving high odometry accuracy without relying on loop closure detection or deep learning methods.

In the context of map-based localization tasks using millimeter-wave radar, Yin et al. \cite{yin2020radar} introduced a method that leverages a prior map generated from LiDAR. Subsequently, Burnett et al. \cite{burnett2022we} compared three topometric localization systems: radar-only, LiDAR-only, and a cross-modal radar-to-LiDAR system and found that the radar-only pipeline achieved competitive accuracy while requiring a significantly smaller map. Additionally, Hong et al. \cite{hong2023large} proposed a localization method using OpenStreetMap (OSM), which eliminated the need for a prior map and combined this approach with a Kalman filter to achieve localization.

\section{METHODOLOGY}

A system overview is depicted in Fig. 2, which presents a weakly supervised learning approach for radar semantic segmentation that leverages corrected LiDAR semantic data as the supervisory signal. In Section III.A, we introduce a pre-processing method that employs sensor field-of-view disparity filtering. In Section III.B, we detail the method for correcting inaccurate labels. Finally, in Section III.C, we showcase how to use the semantic information for various downstream tasks.

\subsection{LiDAR Data Preprocessing}

Conventionally, preprocessing was performed on bird's-eye view (BEV) images generated from the projection of LiDAR point clouds. However, our preprocessing method operates directly on the point cloud data. 

\textbf{FOV Filter:} Building on Lee's method \cite{lee2022patchwork} for removing ground points from LiDAR data, we first align the LiDAR coordinate system with that of the radar. Subsequently, we eliminate the corresponding LiDAR point cloud data based on the differences in the fields of view of the two sensors. As shown in Fig. 3, compared to radar, LiDAR has a wider vertical field of view. To address this, we calculate the angle between the line connecting each LiDAR point and the origin of the new coordinate system with the horizontal plane. Points that fall outside the radar's vertical field of view need to be removed. 
\begin{figure}[thpb]
    \centering
    \includegraphics[width=\linewidth]{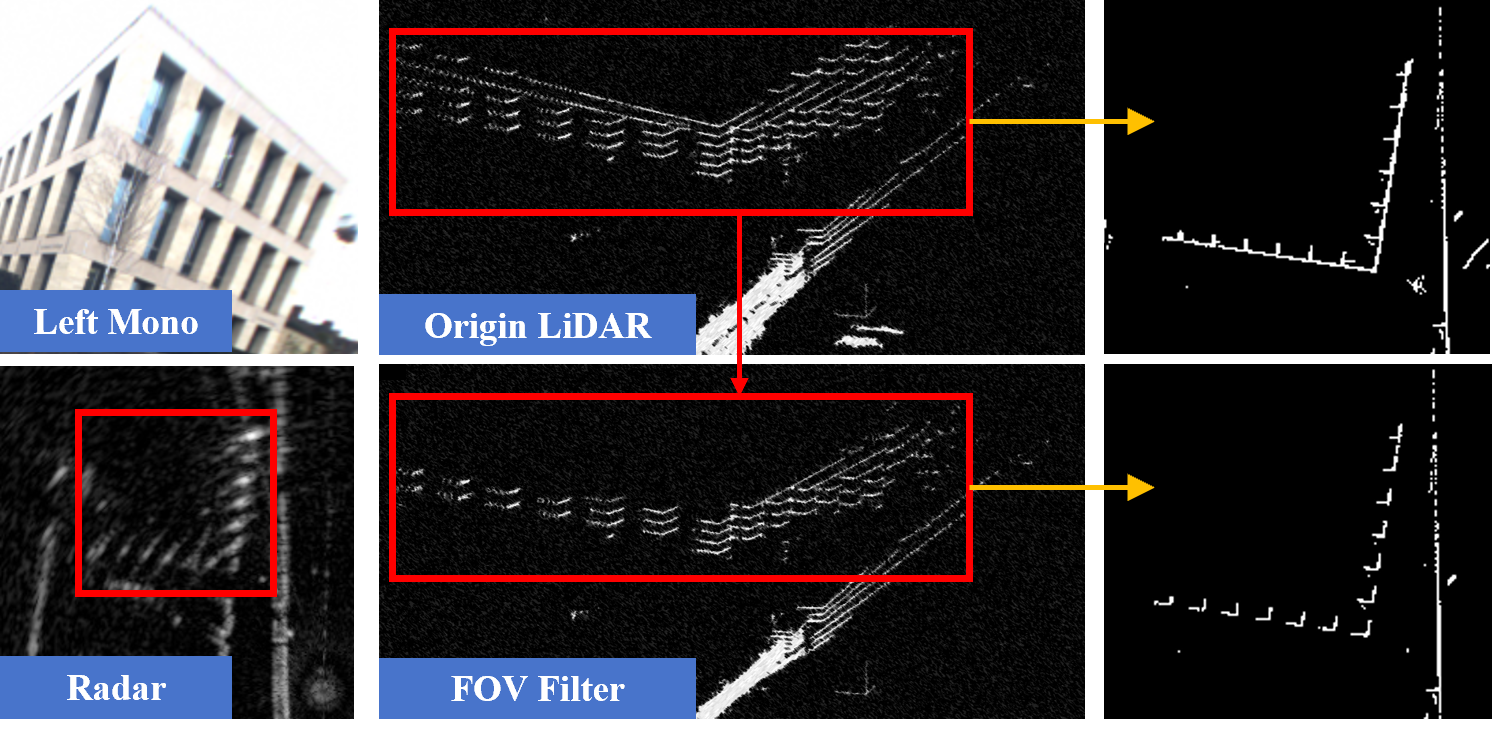}
        \caption{The red box indicates the building captured in the left mono camera image. While the radar can only detect information from the level with windows, the LiDAR can capture both the same level as the radar and the section between the two window levels.}
    \label{FOV_color}
\end{figure}
Fig. 4 shows the results obtained by training with two sets of LiDAR data, one before the FOV filter and one after. The radar images generated using the origin LiDAR data as the supervisory signal contain a large number of false positives (FP). In contrast, when using the preprocessed data for training, the results exhibit more distinct structural features with only very few FP.

\textbf{Projection:} Ambiguity can arise during the projection process because points with different labels may overlap along the height dimension, potentially causing conflicts in the resulting image. To address this, we project three distinct labels separately to generate corresponding BEV maps.
\begin{figure}[thpb]
    \centering
    \includegraphics[width=\linewidth]{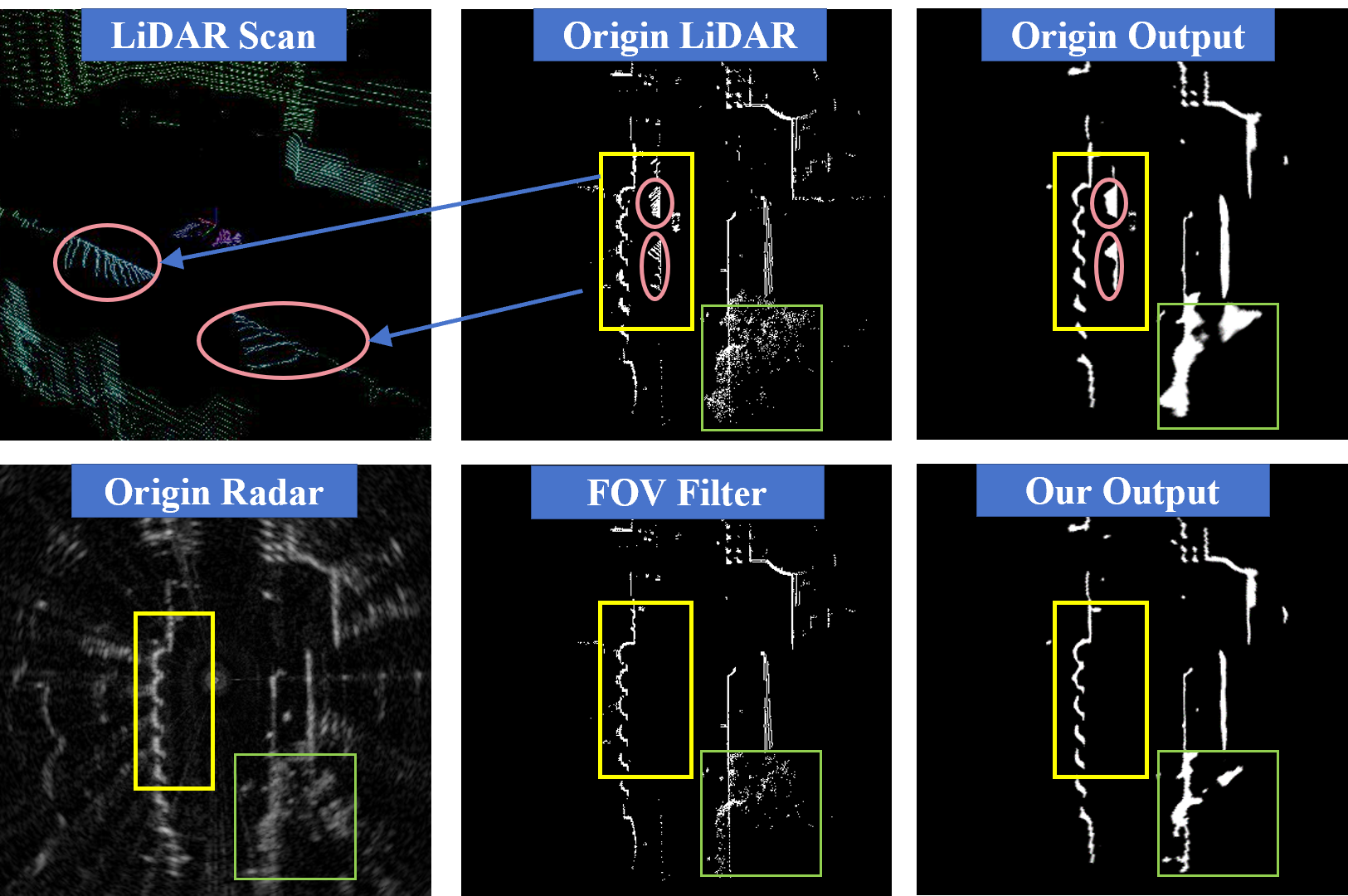}
        \caption{ Comparison before and after the FOV Filter. Colorful rectangles indicate that after the filter, a significant amount of radar-invisible data has been removed from the supervisory LiDAR signal, resulting in a noticeable reduction in false positives in the network's output. Red circles in yellow rectangles highlight that the FOV Filter effectively removes residual ground point clouds. }
    \label{FOV_ground}
\end{figure}

\subsection{Label Refinement} 
We first apply the ground segmentation method to remove the ground point cloud. The remaining point cloud is then fed into the deep learning model to obtain the corresponding semantic labels. To facilitate the refinement of labels, we consolidate all labels into four categories: noise, vehicle, vegetation and building. The noise category primarily includes objects appearing as small clusters in radar points, and pedestrian is classified into noises as segmenting them as an additional class increases the complexity of the semantic segmentation task while also offering limited benefits for downstream tasks. After analyzing the generated point cloud labels, we find that the semantic segmentation of vehicle is highly accurate. However, issues still arise in the segmentation of vegetation and building.

\begin{figure}[t]
    \centering
    \includegraphics[width=\linewidth]{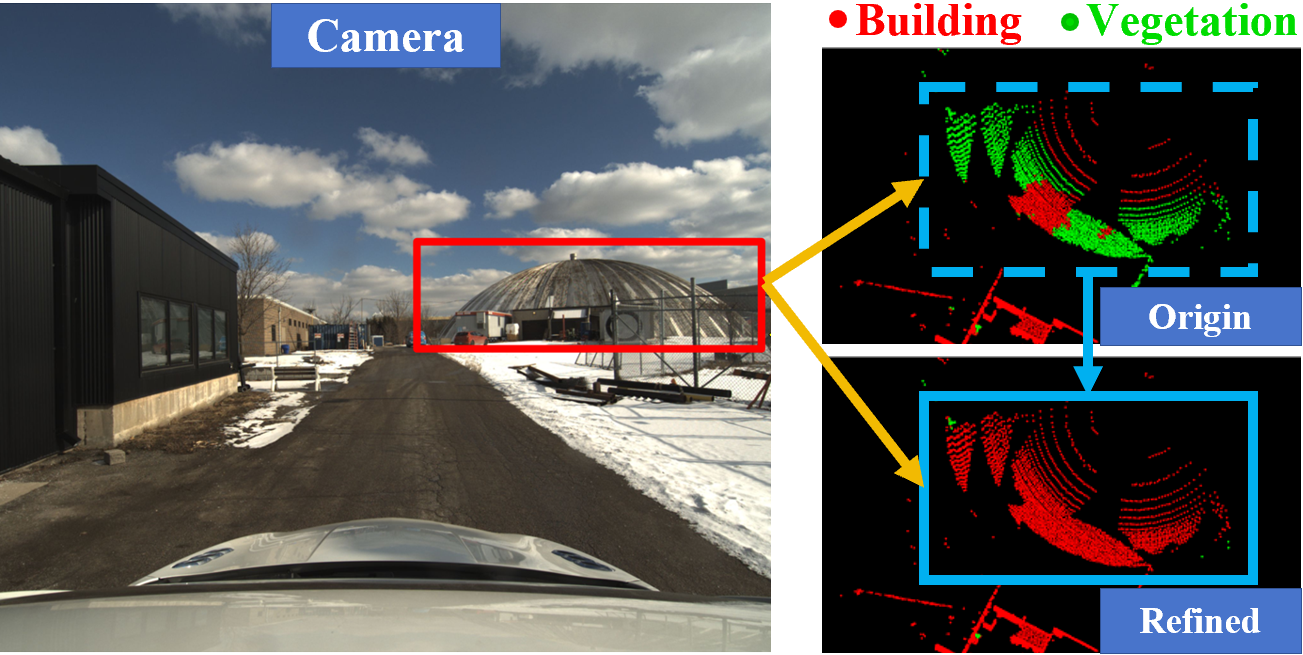}
        \caption{Comparison before and after the label refinement. The red rectangle in the camera view contains a building that has been mistakenly segmented as vegetation. Since the matrix formed by its point cloud coordinates produces two significantly larger singular values (compared to the third), we are able to correct the label to the proper one. }
    \label{SVD}
\end{figure}

As shown in Fig. 5, the first issue is that some buildings are incorrectly segmented as vegetation. To address this, we leverage the differences in the structural features of point clouds generated from building and vegetation after LiDAR scanning. Point clouds formed by building typically approximate a straight line or a plane, whereas point clouds formed by vegetation tend to be concentrated within a specific region and exhibit a more random distribution. We traverse all points labeled as vegetation and apply Singular Value Decomposition (SVD) to obtain three eigenvalues that describe the point cloud's geometric structure. By evaluating these eigenvalues, we determine whether the point cloud formed by the vegetation-labeled points and their neighboring points satisfies the conditions of forming a straight line or a plane. If the conditions are confirmed, we correct the vegetation label to the building label.

As shown in Fig. 6, the second issue is the mixing of building points within the vegetation point cloud, leading to misclassification and reducing segmentation accuracy, which is exacerbated by the model’s difficulty in fully segmenting entire plants. To address this, we implement a method to spatially enclose individual vegetation structures using Axis-Aligned Bounding Boxes (AABB) \cite{AABB}. Specifically, we employ the DBSCAN algorithm \cite{DBSCAN} to group all vegetation-labeled points into multiple clusters. For each cluster with a sufficient number of points, an AABB is calculated to approximate the spatial extent of the corresponding vegetation. All points within the bounding box are then considered as vegetation, allowing us to correct potential labeling errors. Our label refinement method is summarized in Algorithm 1.
\input{code/LabelRefinement}

\subsection{Radar Segmentation and Downstream Tasks}

The refined semantic labels from LiDAR are then used to enhance the radar data, improving the performance in subsequent downstream tasks:

\subsubsection{Radar Odometry}
Our radar odometry is implemented based on CFEAR \cite{adolfsson2022lidar}, which extracts geometrical features from high intensity points without distinguishing high intensity noises and real targets to estimate the 3-degree-of-freedom pose of ground vehicles. Thanks to the building-related semantic information, we can extract feature points that remain stable across consecutive frames. 

Additionally, we incorporate the rotation angles provided by the IMU as a prior, which improves both the rotation and speed of the estimation process.

\begin{figure}[t]
    \centering
    \includegraphics[width=\linewidth]{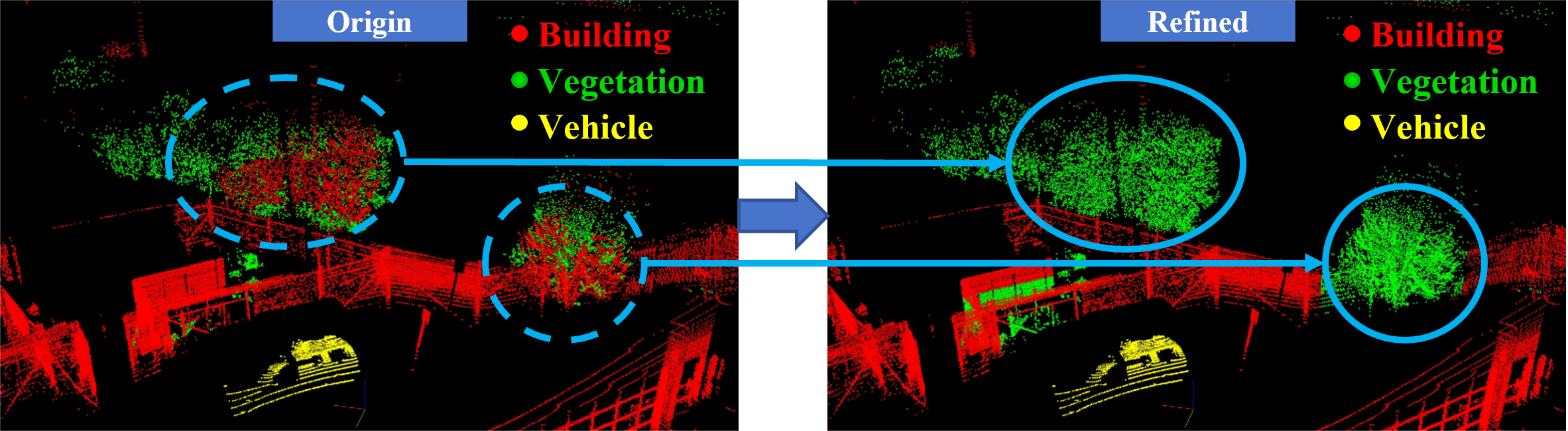}
        \caption{The plant point cloud within the dashed ellipses contains a significant number of mislabeled points. By using the method we propose to enclose their spatial occupancy, all points within this space can be segmented as plants which are shown in the solid-line ellipses. }
    \label{AABB}
\end{figure}
\subsubsection{Localization Using Free Geographic Database}
We use OpenStreetMap (OSM) which contains building information as the map for localization. The radar feature points detected by radar within the LiDAR perception range are relatively sparse, so we utilize a sliding window method \cite{kung2022radar} to perform semantic segmentation on one radar frame that extends beyond the LiDAR's range. This allows us to obtain more information for improving localization. To enhance the stability of registration, we also consider the continuity between consecutive frames. After registering the first few frames, we keep tracking the registered walls and their positions relative to the radar (i.e., left or right). In subsequent frames, we increase the weight of points that correspond to previously registered walls, while decreasing the weight of points with incorrect positional relationships. Fig. 7 illustrates the effectiveness of our method against the origin implementation of \cite{hong2023large}.
\begin{figure}[thpb]
    \centering
    \includegraphics[width=\linewidth]{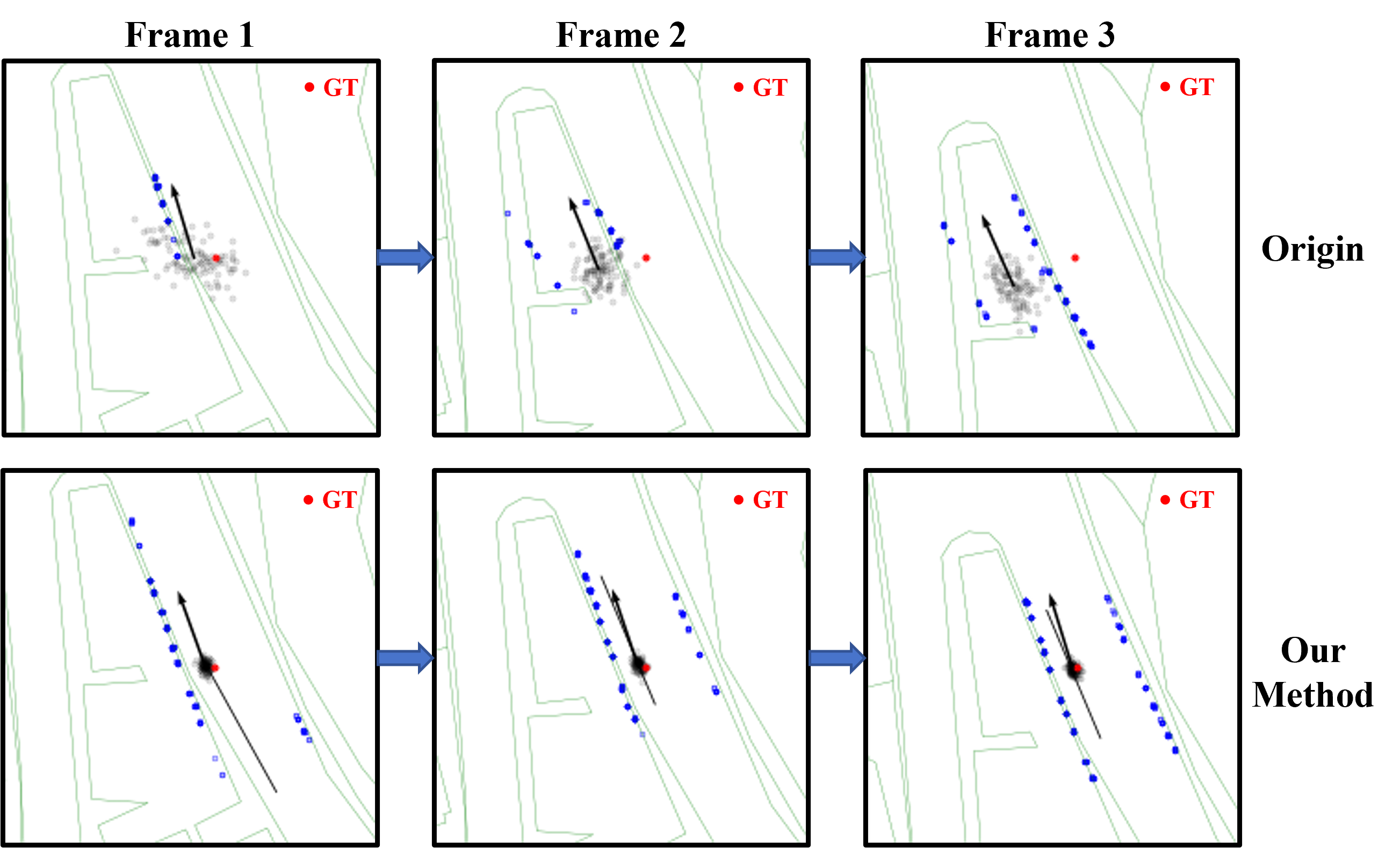}
        \caption{ The blue points represent those that can be successfully registered to wall surfaces. Our method ensures stable registration and localization even when only half of the points are available due to vehicle occlusion. Furthermore, when the other half of the points reappear after the vehicle moves away, they are accurately registered to the corresponding wall surfaces. In contrast, the original method, despite successfully registering a large number of points, often results in incorrect positional relationships. }
    \label{OSM}
\end{figure}

\section{EXPERIMENT SETUP}

\subsection{Datasets}
We conduct tests on two publicly available datasets: the Boreas Dataset \cite{burnett2023boreas} and the Oxford Radar RobotCar Dataset \cite{barnes2020oxford}. The Boreas Dataset provides a Navtech CIR304-H radar, a 128-beam Velodyne Alpha-Prime 3D LiDAR and IMU data. The Oxford Radar RobotCar Dataset provides a Navtech CTS350x radar and two Velodyne HDL32 LiDARs. 

The Boreas Dataset's LiDAR gathers a much larger number of point clouds, providing superior perception range and accuracy compared to the Oxford Radar RobotCar Dataset. 

\subsection{Data Preprocessing}
We utilize Lee's method \cite{lee2022patchwork} to segment the ground in the LiDAR point cloud. For the remaining point cloud, we apply Lai's method \cite{lai2023spherical} to perform semantic segmentation, generating 16 different labels. These labels are then consolidated into four categories: vehicle, vegetation, building, and noise. We further refine and correct these labels using our proposed method. For the first three categories, we project them along the height axis to generate bird's-eye views, which serve as supervisory signals for radar segmentation. These projections are represented in polar coordinates to facilitate the subsequent application of Kung's method \cite{kung2022radar}, enabling the network to utilize raw radar data beyond the perception range of the LiDAR sensor.
\subsection{Training Implementation Details}
We train the network with four Nvidia RTX4090 GPUs and a batch size of 200 over 50 epochs. For the task of radar segmentation, we use a combination of Focal loss \cite{lin2018focal} and Dice loss \cite{milletari2016v}, both with the same weight. Using either loss function alone cannot complete the task: using Focal loss alone would misclassify a large number of noise points as obstacles while using Dice loss alone would fail to recognize any valid radar data. The network of the task mainly follows the U-Net \cite{ronneberger2015u} architecture. We train our model using the AdamW \cite{loshchilov2019decoupled} optimizer with a learning rate of 0.0001, the weight decay of $1\times10^{-6}$, and the momentum of 0.9.

\section{EXPERIMENTAL RESULT}
\subsection{Radar Semantic Segmentation Result}
The semantic segmentation results on both datasets are shown in Table 1. The experiments in this section require sequences that include odometry ground truth and LiDAR data collected under normal weather conditions. For the Oxford Radar RobotCar Dataset, we use sequences specifically intended for odometry testing as the test set, while for the Boreas dataset, we randomly select three sequences for testing. The remaining portions of both datasets are used for training and validation. 

\input{table/radarsemantic}
\begin{figure}[thpb]
    \centering
    \includegraphics[width=\linewidth]{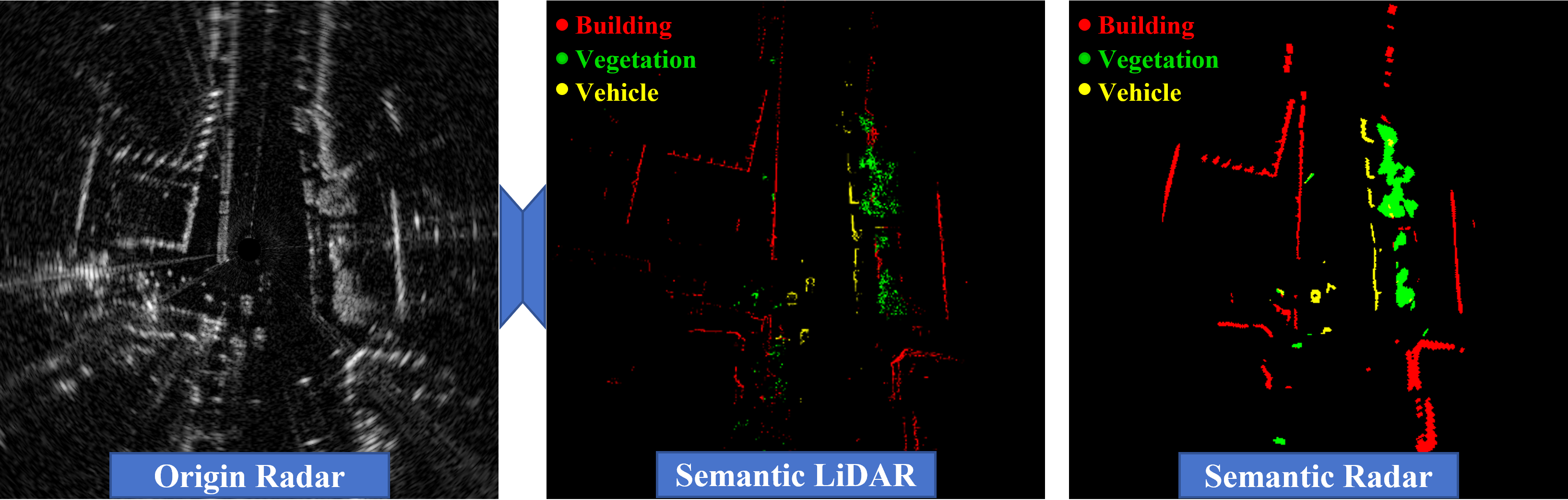}
        \caption{An example of radar semantic segmentation in the Oxford Radar RobotCar Dataset \cite{barnes2020oxford}. }
    \label{RadarSegmentation}
\end{figure}

\input{table/Boreas_odom}

\begin{figure*}[h]
    \centering
    \includegraphics[width=\linewidth]{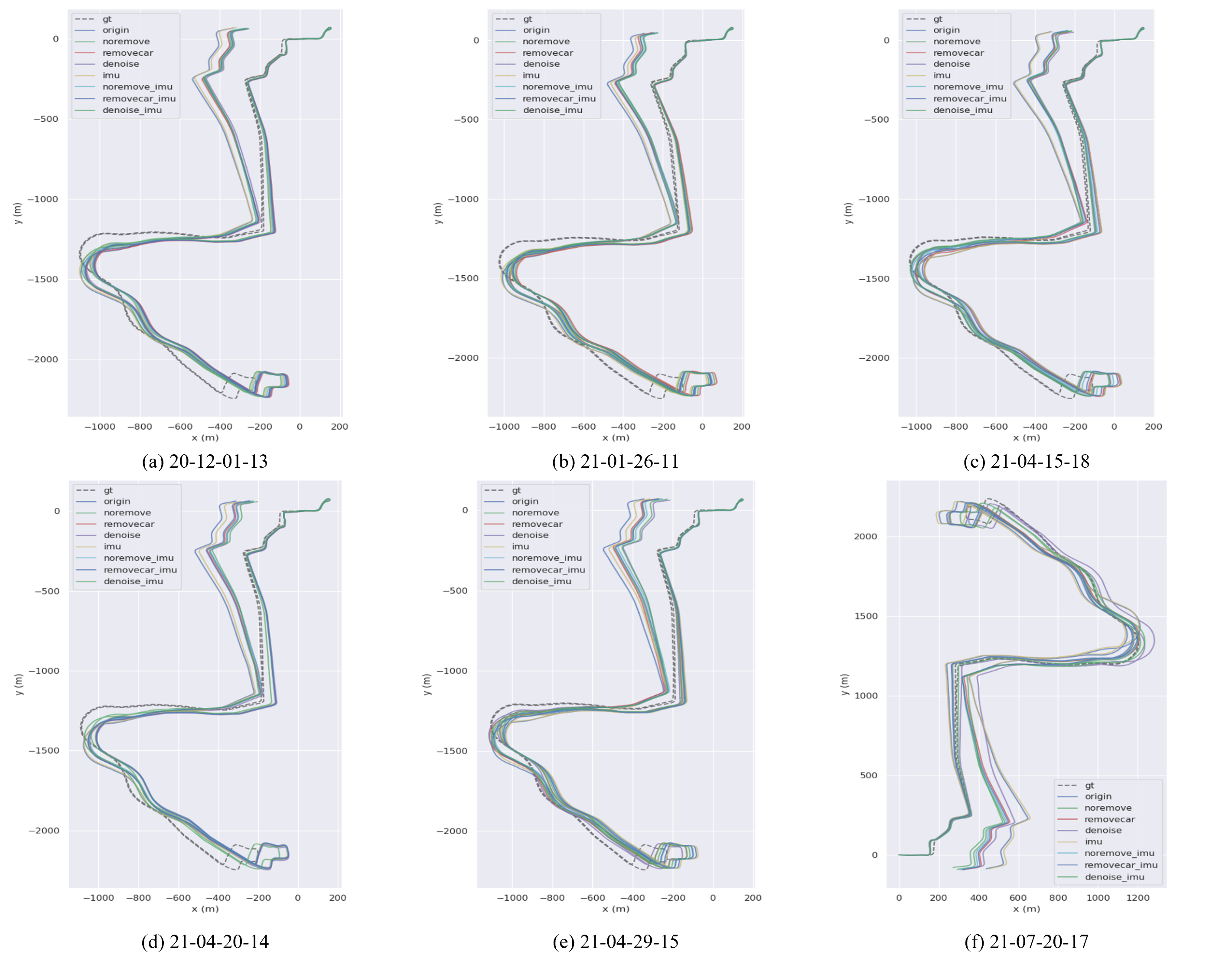}
        \caption{Results of the trajectories for 6 sequences in the Boreas Dataset.}
    \label{odom}
\end{figure*}

The low Intersection over Union (IoU) metric can be attributed to several factors. Firstly, to meet real-time requirements, a lightweight architecture is adopted. However, the shallow depth of convolutional layers limits the model's feature extraction capabilities, resulting in suboptimal segmentation performance. Secondly, due to the inherent differences in the operational principles of radar and LiDAR sensors, their representation of the same scene differs significantly. This discrepancy is particularly pronounced when dealing with objects like trees. LiDAR typically represents trees as discrete point clouds after removing the trunk, while radar tends to display them as a solid mass. The network faces challenges in segmenting these into discrete points, leading to a significant decrease in the IoU metric. 

The IoU metric for the vegetation category differs significantly between the two datasets, with Boreas results being much higher than Oxford's, primarily due to differences in the data collection environments. The Boreas Dataset was collected mainly in suburban areas, where the routes contain a large number of trees, and buildings are more dispersed, making it easier for the network to segment the vegetation. In contrast, the Oxford Dataset was collected in the densely built-up Oxford city area, where vegetation and building are intricately interwoven, making the distinction between them considerably more challenging. An example of radar semantic segmentation is visualized in Fig. 8.

\subsection{Radar Odometry Result}
On the Boreas Dataset, we use CFEAR-3 \cite{adolfsson2022lidar} with 10 keyframes as the baseline. As an ablation study, we conduct comparative experiments with different semantic classes removed: 1). none removed; 2). vehicle removed; and 3). only building (both the vehicle and vegetation are removed), in combination with and without IMU. As shown in Table \ref{Boreas_odom}, in most sequences (5 out of 6), the method with both vehicle and vegetation removed and with IMU achieves the best performance in terms of translation errors. The major improvement is attributed to two reasons: 1). the removal of moving cars and trees facilitates more stable feature registration; and 2). the introduced gyroscope rotation information provides a good motion prior during sharp turns. Moreover, in all sequences, the odometry accuracy achieved using only our semantic data surpasses the results obtained by solely incorporating IMU data. This further demonstrates the reliability and effectiveness of our approach. The estimated trajectories are shown in Fig. 9.

There is significant variation in the translation accuracy improvement across different sequences. Since the primary goal of our model is to remove radar noise, we analyze the images obtained after applying the k-strongest filter \cite{adolfsson2022lidar} to both the raw data and the data enhanced by our method. To assess the difference between the two images, we use the Mean Squared Error (MSE), which calculates the average of the squared differences between the pixel values of the two images. A smaller MSE indicates higher similarity between the images. The formula for MSE is as follows:
$$\mathrm{MSE}=\frac1{mn}\sum_{i=1}^m\sum_{j=1}^n[I_1(i,j)-I_2(i,j)]^2$$
\( I_1(i,j) \) and \( I_2(i,j) \) represent the pixel values of the two images at the position \((i,j)\), while \( m \) and \( n \) denote the height and width of the images, respectively. Fig. 10 shows the MSE values and the percentage
improvement in translation accuracy across the sequences in Table \ref{Boreas_odom}. A larger MSE value indicates a more significant difference between the two images, suggesting that the original data, when processed using the k-strongest filter, contains a substantial amount of noise. Consequently, the data enhanced by our method leads to a greater improvement in odometry translation accuracy.
\begin{figure}[thpb]
    \centering
    \includegraphics[width=\linewidth]{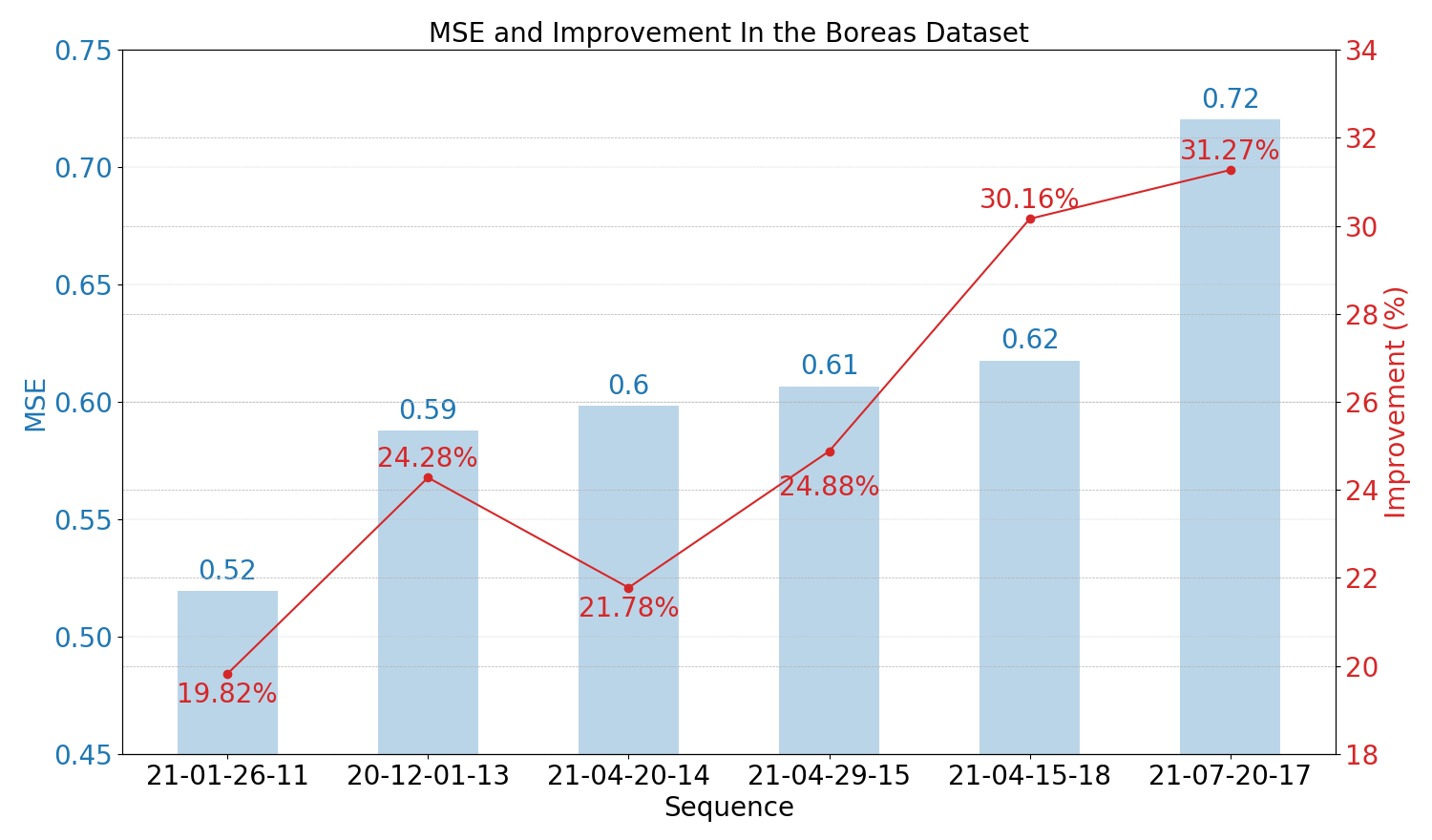}
        \caption{The plot shows the MSE of the difference between the raw data and the data enhanced by our method and the corresponding improvement in odometry translation accuracy across different sequences from the Boreas Dataset. The blue bars represent the MSE values, while the red line shows the percentage improvement in translation accuracy.}
    \label{MSE_grid}
\end{figure}

In the ICRA 2024 Radar in Robotics odometry competition, the ground truth poses of the test sequences have been withheld. Submissions are ranked based on their translational drift [\%] on the benchmark. We use the method that achieved the best results in Table \ref{Boreas_odom}. 
In particular, during the test phase of this competition, the sequence 2021-09-09-15-28 was collected in an urban environment, which differs from all other sequences. Despite this difference, our method is still able to achieve satisfactory results on this sequence, further demonstrating the robustness and generalization of our approach.

On the Oxford Radar RobotCar Dataset, we use CFEAR-3 \cite{adolfsson2022lidar} with origin parameters as the baseline. Due to the absence of actual IMU data, our discussion is limited to three cases: 1). none removed; 2). Vehicle removed; and 3). Only building (both the vehicle and vegetation are removed). As shown in Table \ref{Oxford_odom}, in the last three sequences, the method with only building removed achieves the best performance in terms of translation errors.

\input{table/Oxford_odom}

\subsection{Localization Using Free Geographic Database}
Since our method significantly improves odometry accuracy on the two datasets, we conduct the following evaluation for localization. First, we replace the original radar data with data generated by our model, which contains only building semantic information. Next, we substitute Hong's odometry \cite{hong2023large} with the odometry generated by our method in Section V-B and conduct the tests. In both of the above tests, we use our newly proposed method, which considers the continuity between consecutive frames.  As shown in Table \ref{table5},
\begin{table}[h!]
\centering
\begin{threeparttable}        
\caption{Average Position Error (in Meters)}
\label{table5}
\begin{tabular}{|c|*{4}{c|}*{4}{c|}}
  \hline
  \multirow{2}{*}{Method} & \multicolumn{4}{c|}{Boreas\tnote{1}} & \multicolumn{4}{c|}{Oxford\tnote{2}}  
  \\
  \cline{2-9} & 01 & 02 & 03 & 04 & 01 & 02 & 03 & 04 \\ 
  \hline
  Hong \cite{hong2023large} & 4.7 & 4.9 & 4.9 & 3.8 & 5.4 & 3.4 & 4.1 & 4.1  \\
  \hline
  Semantic Data  & 3.8 & 4.0 & 4.1 & 3.9 & 5.0 & 3.6 & 3.6 & 3.5 \\
  Our Odometry\tnote{3} & \textbf{3.0} & \textbf{3.0} & \textbf{3.4} & \textbf{3.7} & \textbf{4.5} & \textbf{3.3} & \textbf{3.3} & \textbf{3.4} \\
  \hline
\end{tabular}
\begin{tablenotes}    
        \footnotesize               
        \item[1]  01: 11-02-11-16, 02: 11-14-09-47, 03: 11-16-14-10, 04: 11-23-14-27. 
        \item[2] 01: 10-11-46-21, 02: 10-12-32-52, 03: 18-14-46-59, 04: 18-15-20-12.        
        \item[3] Our Odometry contains Semantic Data. 
      \end{tablenotes}            
\end{threeparttable} 
\end{table}
in all sequences, by using our data and odometry, we achieve the best performance in terms of average position error. On the Boreas Dataset, the first three sequences all showed at least a 30\% improvement, while the fourth sequence has a smaller gain. And on the Oxford Robotcar Dataset, all sequences have little improvement. This is primarily because Hong's method \cite{hong2023large} performs well on these sequences, with the odometry already achieving high accuracy.
\section{CONCLUSION}
In this paper, we propose a weakly supervised learning method for the semantic segmentation of radar. A label refinement scheme is proposed to refine the LiDAR semantic labels, which is further used in a cross-modal supervision manner to train the radar semantic segmentation model. Applying the semantic information generated by our model to the downstream tasks of odometry and localization significantly improves the estimation accuracy, demonstrating the effectiveness of our approach.

\normalem
\bibliographystyle{ieeetran}
\bibliography{ref}

\end{document}

%% file: code/LabelRefinement.tex
\begin{algorithm}
\renewcommand{\algorithmicrequire}{\textbf{Input:}}
\renewcommand{\algorithmicensure}{\textbf{Output:}}
\caption{Label Refinement}
    \begin{algorithmic}[1]
    \REQUIRE{Point cloud with labels: $\begin{aligned}A=\{a_{label}\}\end{aligned}$;%\\
    }
        \STATE \textbf{Parameters:} search radius $r$, minimum points $n$;
        \STATE {Clusters $C\leftarrow\text{DBSCAN}(A_{vegetation})$;}
        \FOR {$c_i\in C$}
        \STATE {$aabb\leftarrow\text{ComputeAABB}(c_i)$;}
        \STATE {$A\leftarrow\text{RefineVegetationPoints}(A,aabb)$;}
        \ENDFOR
        \FOR {$p_j\in A_{vegetation}$}
        \STATE {$P, n_P\leftarrow\text{SearchNearPoints}(A,p_j,r)$}
        \IF { $n_P>n$}
            \STATE {$\sigma_1,\sigma_2,\sigma_3\leftarrow\text{SVD}(P)$;}
            \STATE {$Structure\leftarrow\text{AccessStructure}(\sigma_1,\sigma_2,\sigma_3)$;}
            \IF{$Structure==\text{``Line"}\; or \; \text{``Plane"}$}
            \STATE {$A\leftarrow\text{RefineBuildingPoints}(A,P)$;}
            \ENDIF
        \ENDIF
        \ENDFOR
    \ENSURE{Point cloud with refined labels $A$;}
    \end{algorithmic} 
\end{algorithm}

%% file: table/radarsemantic.tex
\begin{table}[h!]
\centering
\caption{Radar semantic segmentation results (Metric: IoU)}
\label{table1}
\begin{tabular}{|l|c|c|c|}
\hline
                & building & vehicle     & vegetation    \\
\hline
The Oxford Dataset\cite{barnes2020oxford} & 0.398 & 0.231 & 0.297 \\
The Boreas Dataset\cite{burnett2023boreas}                & 0.340 & 0.200 & 0.425 \\
\hline
\end{tabular}
\end{table}

%% file: table/Boreas_odom.tex
\begin{table*}
  \caption{Evaluation in the Boreas dataset. Drift is measured by the translation error $[\%]$, rotation error (degree/100m). For each sequence, The least translation error and rotation error of each sequence is highlighted in bold.}
  \begin{center}
  \begin{tabular}{ccccccccccccc}
    \toprule
    {\multirow{2}{*}{Methods}} & \multicolumn{2}{c}{20-12-01-13} & \multicolumn{2}{c}{21-01-26-11} & \multicolumn{2}{c}{21-04-15-18} & \multicolumn{2}{c}{21-04-20-14} & \multicolumn{2}{c}{21-04-29-15} & \multicolumn{2}{c}{21-07-20-17}  \\
    \cmidrule(lr){2-3} \cmidrule(lr){4-5} \cmidrule(lr){6-7} \cmidrule(lr){8-9} \cmidrule(lr){10-11} \cmidrule(lr){12-13} 
    & $Trans.$ & $Rot.$ & $Trans.$ & $Rot.$ & $Trans.$ & $Rot.$ & $Trans.$ & $Rot.$ & $Trans.$ & $Rot.$ & $Trans.$ & $Rot.$  \\
    \midrule
     CFEAR-3\cite{adolfsson2022lidar}    &  0.902 & 0.291 & 0.797 & 0.289  &  0.902 & 0.281 & 0.845 & 0.282 & 0.860 & 0.268 & 0.937 & 0.227  \\
    \midrule
   None removed & 0.686&0.283 & 0.656&0.287 & \textbf{0.575}&0.264 & 0.611&0.272 & 0.651&0.257 & 0.624&0.194 \\
Vehicle removed & 0.693&0.285 & 0.655&0.287 & 0.621&0.271& 0.622&0.276 & 0.638&0.256& 0.626&0.198 \\
Only building & 0.683&0.284 & 0.639&0.280 & 0.630&0.267 & 0.661&0.278 & 0.646&0.267 & 0.644&0.211  \\
    \midrule
     IMU & 0.866&0.284 & 0.735&0.279 & 0.886&0.278 & 0.801&0.274 & 0.825&0.260 & 0.966&0.233 \\
None removed + IMU& 0.640&0.276& 0.602&0.279& 0.600&0.263& 0.583&0.267& 0.614&\textbf{0.248}& 0.652&0.199\\
Vehicle removed+ IMU&0.640&0.276&0.594&0.277&0.618&0.267
&0.601&0.270 &0.610&0.250&0.643&0.199\\
Only building + IMU & \textbf{0.638}&\textbf{0.276}& \textbf{0.584}&\textbf{0.272}& 0.605&\textbf{0.261}& \textbf{0.541}&\textbf{0.258}& \textbf{0.592}&0.249& \textbf{0.599}&\textbf{0.191}\\
    \bottomrule
  \end{tabular}
  \end{center}
\label{Boreas_odom}
\end{table*}

%% file: table/Oxford_odom.tex
\begin{table}[h!]
  \caption{Evaluation in the Oxford Radar RobotCar Dataset. %Drift is measured by the translation error $[\%]$, rotation error (degree/100m). For each sequence, The least translation error and rotation error of each sequence is highlighted in bold.
  }
  \setlength{\tabcolsep}{1.75pt}
  \begin{tabular}{ccccccccc}
    \toprule
    {\multirow{2}{*}{Methods}} & \multicolumn{2}{c}{10-11-46} & \multicolumn{2}{c}{10-12-32} & \multicolumn{2}{c}{16-11-53} & \multicolumn{2}{c}{16-13-09}  \\
    \cmidrule(lr){2-3} \cmidrule(lr){4-5} \cmidrule(lr){6-7} \cmidrule(lr){8-9}
    & $Trans.$ & $Rot.$ & $Trans.$ & $Rot.$ & $Trans.$ & $Rot.$ & $Trans.$ & $Rot.$   \\
    \midrule
     CFEAR-3\cite{adolfsson2022lidar}    &  1.26 & \textbf{0.39} & 1.23 & 0.36  &  1.42 & \textbf{0.39} & 1.25 & \textbf{0.39} \\
    \midrule
   None removed  & 1.20&0.41 & 1.06&0.36 & 1.39&0.45 & 1.19&0.40\\
Vehicle removed & \textbf{1.17}&0.41 & 1.07&0.36 & 1.29&0.41 & 1.18&0.41 \\
Only building & 1.19&0.42 & \textbf{1.04}&\textbf{0.36} & \textbf{1.28}&0.41 & \textbf{1.17}&0.41 \\
    \bottomrule
  \end{tabular}
\label{Oxford_odom}
\end{table}